\documentclass[journal,twoside,web]{ieeecolor}
\usepackage{generic}
\usepackage{cite}
\usepackage{amsmath,amssymb,amsfonts}
\usepackage{algorithmic}
\usepackage{graphicx}
\usepackage{algorithm,algorithmic}
\usepackage{hyperref}
\hypersetup{hidelinks=true}
\usepackage{textcomp}
\usepackage{booktabs} 

\usepackage{amsmath}
\usepackage{multirow}
\usepackage{algorithm}
\usepackage{algorithmic}

\usepackage{orcidlink}

\def\BibTeX{{\rm B\kern-.05em{\sc i\kern-.025em b}\kern-.08em
    T\kern-.1667em\lower.7ex\hbox{E}\kern-.125emX}}
\begin{document}
\title{BEAT-Net: Injecting Biomimetic Spatio-Temporal Priors for Interpretable ECG Diagnosis}

\author{Runze Ma \orcidlink{0009-0001-2620-0575}, Haonan Lyu \orcidlink{0009-0005-2932-816X}, Shunbo Jia \orcidlink{0009-0003-9854-9930}, Qiang Yang, \textit{Member, IEEE} \orcidlink{0000-0002-5202-7892}, Muzi Xu, \textit{Graduate Student Member, IEEE} \orcidlink{0000-0001-6381-9863}, Jiaqi Zhang \orcidlink{0009-0006-9479-5706}, Zihe Luo \orcidlink{0009-0004-6434-5582}, and Caizhi Liao \orcidlink{0000-0003-1271-3027}
\thanks{This work has been submitted to the IEEE for possible publication. Copyright may be transferred without notice, after which this version may no longer be accessible.}
\thanks{Corresponding authors: Qiang Yang and Caizhi Liao.}
\thanks{Runze Ma is with Department of Bioelectronics, Faculty of Biomedical Engineering, Shenzhen University of Advanced Technology, Shenzhen 51810, China, and also with School of Information Technology, Monash University Malaysia, Subang Jaya 47500, Malaysia, e-mail: rmaa0033@student.monash.edu.}
\thanks{Haonan Lyu and Caizhi Liao are with Department of Bioelectronics, Faculty of Biomedical Engineering, Shenzhen University of Advanced Technology, Shenzhen 51810, China, e-mail: SUAT25060153@stu.suat-sz.edu.cn; liaocaizhi@suat-sz.edu.cn.}
\thanks{Shunbo Jia is with Faculty of Innovation Engineering, Macau University of Science and Technology, Macau, China, e-mail: 2240003657@student.must.edu.mo.}
\thanks{Qiang Yang is with Department of Computer Science and Technology, University of Cambridge, Cambridge CB2 1TN, U.K., e-mail: qy258@cam.ac.uk.}
\thanks{Muzi Xu is with Electrical Engineering Division, Department of Engineering, University of Cambridge, Cambridge CB3 0FA, U.K., e-mail: mx247@cantab.ac.uk.}
\thanks{Jiaqi Zhang is with Department of Electrical and Electronic Engineering, The University of Hong Kong, Hong Kong SAR, China, e-mail: zhangjiaqi@connect.hku.hk.}
\thanks{Zihe Luo is with Institute of Health Informatics, University College London, London WC1E 6BT, U.K., e-mail: zephyr@hhu.edu.cn.}}

\maketitle

\begin{abstract} 
Automated electrocardiogram diagnosis using deep learning remains limited by signal-agnostic representations that treat multi-lead recordings as undifferentiated time-series or images, forcing models to rediscover physiological structure implicitly. This leads to data inefficiency, poor generalization, and opaque decision boundaries misaligned with clinical reasoning. We present BEAT-Net, a supervised biomimetic framework that integrates QRS-centered biological tokenization with a hierarchical architecture mirroring the cardiologist's workflow. A QRS tokenizer converts continuous signals into semantically complete heartbeat sequences, which are processed through four specialized stages: morphological feature extraction via a Word Encoder, lead-invariant normalization through a Spatial Operator, temporal context injection by a Temporal Operator, and global reasoning using a Transformer-based Sentence Encoder. Evaluated across three large-scale benchmarks including PTB-XL, CPSC2018, and CSN, BEAT-Net achieves diagnostic accuracy of 0.924 AUC, comparable to dominant CNN baselines at 0.925 AUC, while reducing parameters by 95 percent from 2.06 million to 0.7 million. Critically, BEAT-Net surpasses the 39.5-million-parameter foundation model HeartLang on morphological Form classification, reaching 0.901 AUC compared to HeartLang's 0.832 AUC, while attaining full CNN-level performance using only 35 percent of training data and exhibiting superior cross-dataset generalization. Learned attention patterns spontaneously align with established clinical heuristics, demonstrating that explicit physiological structure provides a more efficient and interpretable alternative to massive pre-training for clinical deployment.
\end{abstract}


\section{Introduction}

Cardiovascular diseases (CVD) remain the leading cause of global mortality~\cite{tsao2023heart}, and the 12-lead electrocardiogram (ECG) serves as the clinical gold standard for diagnosing cardiac anomalies~\cite{writing20212021}. Automated ECG analysis via artificial intelligence has become central to supporting physician interpretation and managing growing clinical workloads. Despite substantial progress with Convolutional Neural Networks (CNNs) and Vision Transformers (ViTs)~\cite{dosovitskiy2020image}, the dominant supervised paradigm treats multi-lead recordings as undifferentiated one-dimensional time-series or two-dimensional images~\cite{ansari2025survey}. This signal-agnostic view forces models to rediscover physiological structures such as P-waves, QRS complexes, and T-waves without explicit priors~\cite{hong2019mina,lan2022intra}, requiring enormous amounts of annotated data to learn patterns that cardiologists identify through established diagnostic principles. These models also remain opaque black boxes~\cite{rudin2019stop}, with decisions that often misalign with the hierarchical reasoning used in clinical practice.

The fundamental challenge is that signal-agnostic methods implicitly reinvent centuries of accumulated knowledge about cardiac physiology. A trained cardiologist identifies QRS complexes to assess individual beat integrity, mentally normalizes morphological variations across lead perspectives rooted in anatomical orientation, and synthesizes temporal rhythmicity to detect arrhythmias. Deep learning models starting from raw waveforms must learn all these concepts from statistical correlations alone, resulting in severe data inefficiency and poor generalization across datasets.

Recent advances in Self-Supervised Learning (SSL) and ECG Foundation Models address part of this problem by recasting ECG as a language and using tokenization to model long-range dependencies~\cite{weng2025self,jin2025reading}. These approaches achieve strong semantic representations and have demonstrated improved generalization over signal-agnostic CNNs in pre-training settings. However, the computational expense of billion-parameter architectures and discrete Vector Quantization makes deployment in resource-constrained clinical settings impractical~\cite{han2024ecg,an2024research,spicher2021edge}, while frozen feature extractors fail to adapt to task-specific nuances even with increasing supervision. This creates a critical gap between methods that are either data-inefficient or computationally prohibitive, with no solution that achieves parameter efficiency, data efficiency, and interpretability simultaneously.

We propose that the semantic benefits of tokenization can be retained within a lightweight supervised framework grounded in clinical principles. We introduce BEAT-Net (\textbf{B}iomimetic \textbf{E}CG \textbf{A}nalysis with \textbf{T}okenization), a supervised framework that operationalizes the cardiologist's diagnostic workflow into learnable architectural components~\cite{goldberger2018goldberger}. Rather than treating ECG as undifferentiated signal, BEAT-Net explicitly integrates QRS-centered tokenization, lead-specific normalization, and temporal rhythm modeling as fundamental design elements. This biomimetic approach allows the model to focus learning on pathological discrimination rather than rediscovering basic physiology.

We adopt a QRS-tokenization strategy~\cite{jin2025reading} that converts continuous signals into biologically aligned heartbeat tokens, where each token preserves the P-QRS-T complex and represents a complete cardiac cycle. These tokens pass through a modular four-stage pipeline that mirrors clinical reasoning. A Word Encoder extracts beat-level morphology using residual blocks~\cite{he2016deep}, a Spatial Operator normalizes representations across lead perspectives, a Temporal Operator reconstructs rhythmic sequences, and a Sentence Encoder synthesizes global patterns through attention~\cite{vaswani2017attention}. This explicit hierarchy anchors all intermediate representations in physiological meaning while enabling efficient end-to-end learning.

Our main contributions are as follows.

\begin{itemize} 
    \item \textbf{Supervised Biomimetic Paradigm.} We integrate QRS-centered biological tokenization into supervised learning. This approach aligns each token with a complete cardiac cycle, preserves the P-QRS-T complex, and avoids arbitrary patching used in signal-agnostic ViT style methods. The result combines semantic structure with low computational overhead.

    \item \textbf{Biomimetic Hierarchical Architecture.} BEAT-Net mirrors the clinical workflow through morphological, spatial, and temporal operators. This hierarchy preserves waveform detail while resolving rhythm dependencies, improving interpretability by grounding each stage in physiological structure.

    \item \textbf{Robustness, Efficiency, and Interpretability.} Experiments on PTB-XL~\cite{wagner2020ptb}, CPSC2018~\cite{liu2018open}, and CSN~\cite{zheng2020optimal,zheng2022large} show that BEAT-Net matches leading 1D-CNN baselines~\cite{strodthoff2020deep} with only 0.7 million parameters versus 2.06 million for the baseline. Compared with HeartLang+MLP, BEAT-Net delivers stronger morphological performance and better cross dataset generalization while using far fewer parameters. The model reaches full performance using 35\% of annotated data and learns attention patterns that reflect clinical heuristics for rhythm~\cite{sandau2017update} and morphology~\cite{sokolow1949ventricular}, while downweighting redundant leads to reflect physiological constraints.
\end{itemize}

\section{Related Work}

\subsection{Supervised Deep Learning for ECG Analysis}

The application of deep learning to ECG analysis has historically been dominated by supervised paradigms that treat physiological signals as undifferentiated time-series data~\cite{jaya2025advances, petmezas2022state}. Early innovations, such as the Adaptive 1D-CNNs by Kiranyaz et al., established the end-to-end learning philosophy~\cite{kiranyaz2015real}. These models demonstrated that convolutional kernels could serve as adaptive feature extractors to capture morphological patterns without manual feature engineering~\cite{hannun2019cardiologist, kiranyaz2015convolutional}. Subsequent architectures like ResNets introduced residual connections to enable the training of deep networks~\cite{jing2021ecg, yang2025robust}. These models typically aggregate features across 12 leads via global pooling~\cite{wasimuddin2021multiclass}. Meanwhile, InceptionTime addressed the multi-scale nature of cardiac pathology through parallel convolutions of varying lengths~\cite{crocker2021inceptiontime, cao2023incepse}, which can simultaneously capture high-frequency QRS complexes and low-frequency T-waves~\cite{zhao2025mak}.
However, these methods rely on implicit physiological modeling. By processing the ECG as a raw tensor of voltage values, these models attempt to rediscover basic physiological laws, including the definition of a heartbeat and the spatial orientation of leads, solely through statistical correlations~\cite{bodini2021opening, duran2021afraid}. While such implicit learning can eventually succeed, this signal-agnostic approach incurs significant data inefficiency, requiring massive annotated datasets to compensate for the absence of structural priors that are already well-defined in clinical medicine.

\subsection{Self-Supervised Learning and ECG Foundation Models}

To address annotation bottlenecks, the field has shifted toward Self-Supervised Learning (SSL)~\cite{ronan2024brugada}. Early contrastive frameworks like CLOCS leveraged physiological invariances, including spatial consistency across leads and temporal consistency within patients, to learn robust representations~\cite{kiyasseh2021clocs}. Recently, this has evolved into ECG Foundation Models that reconceptualize the ECG as a language~\cite{mousavi2021ecg, mckeen2025ecg}. A central mechanism in this paradigm is Tokenization, which diverges into two strategies. Signal-agnostic methods slice signals into fixed-duration patches~\cite{wang2024anyecg, na2024guiding}. For instance, AnyECG and ST-MEM adopt a Vision Transformer style approach but often sever P-waves or QRS complexes across boundaries~\cite{shivashankara2024ecg}. Conversely, biomimetic approaches like HeartLang employ QRS-centered tokenization to discretize signals into biologically meaningful heartbeat units~\cite{jin2025reading}. While these models achieve remarkable semantic understanding, they typically rely on billion-parameter architectures and discrete Vector Quantization (VQ)~\cite{han2024ecg}, imposing prohibitive computational costs that hinder deployment in resource-constrained clinical settings~\cite{an2024research, spicher2021edge}.
Parallel to the rise of large models, some works have attempted to inject domain knowledge into smaller models. For instance, MINA proposed multilevel knowledge-guided attention to weight specific signal regions~\cite{hong2019mina}. However, such methods often rely on complex feature engineering branches rather than fundamentally restructuring the input data~\cite{najia2025enhanced}.

\subsection{Knowledge-Guided Architectural Priors}

While previous works such as MINA~\cite{hong2019mina} attempt to inject domain knowledge via auxiliary attention branches, they often treat ECG as a generic waveform, failing to fundamentally align the model's internal logic with the actual diagnostic workflow. Our proposed BEAT-Net bridges this gap by restructuring the learning process into a biomimetic hierarchical framework that operationalizes the cardiologist's cognitive heuristic into distinct architectural constraints.

Analogous to the initial clinical phase where a physician identifies QRS complexes to evaluate individual beat integrity and screen for local pathologies like hypertrophy or ischemia, we employ Biological Tokenization as a structural prior. This design forces the model to treat the physiological heartbeat, rather than arbitrary sliding windows, as its fundamental semantic unit. Subsequently, to address the spatial complexity where clinicians must mentally normalize morphological variations across different anatomical leads~\cite{de2021clinical}, we embed a Spatial Operator that calibrates feature representations based on lead identity, ensuring view-invariant understanding. Furthermore, mirroring the synthesis of long-term rhythmicity essential for arrhythmia diagnosis~\cite{wagner2009aha}, the framework incorporates a Temporal Operator and a Transformer backend to resolve sequential dependencies across the entire recording.

Unlike foundation models that utilize tokenization primarily for expensive generative pre-training, BEAT-Net leverages this clinically aligned decomposition to achieve efficient supervised learning. This approach addresses the data inefficiency of raw-signal CNNs while maintaining computational tractability compared to large-scale foundation models, offering a framework grounded in physiological interpretability~\cite{hammer2025explainable}.

\section{Method}

Let a multi-lead ECG recording be denoted as $X \in \mathbb{R}^{C \times T}$ where $C$ represents the number of leads and $T$ is the signal duration. Unlike prevalent paradigms that treat $X$ as a generic continuous waveform or image~\cite{ansari2025survey}, we formulate the diagnostic process as a biomimetic probabilistic modeling task that mirrors the hierarchical reasoning of a cardiologist. We define a transformation $\Phi$ that maps the raw signal into a sequence of structured heartbeat tokens $\mathcal{H} = \{h_1, h_2, \dots, h_S\}$. This formulation effectively treats individual heartbeats as fundamental semantic units within a structured physiological sequence, where each token encapsulates a localized cardiac event. Following this, we model the conditional probability $P(y | \mathcal{H})$ using a modular architecture that mimics the linguistic transition from vocabulary to syntax by extracting local morphology via Word Encoder, normalizing view variations via Spatial Operator, and integrating rhythmic context via Temporal Operator and Sentence Encoder. This explicit decomposition enhances data efficiency and ensures interpretability by grounding predictions in distinct physiological components.

\subsection{Biological Tokenization}

In clinical practice, the diagnostic heuristic begins with morphological isolation, where a physician identifies QRS complexes to evaluate the integrity of individual heartbeats, such as assessing QRS width~\cite{hampton2024ecg}. To replicate this, we transform continuous signals into discrete semantic sequences using a QRS-Tokenizer strategy~\cite{jin2025reading}.

\begin{figure}[ht]
\centering
\includegraphics[width=\linewidth]{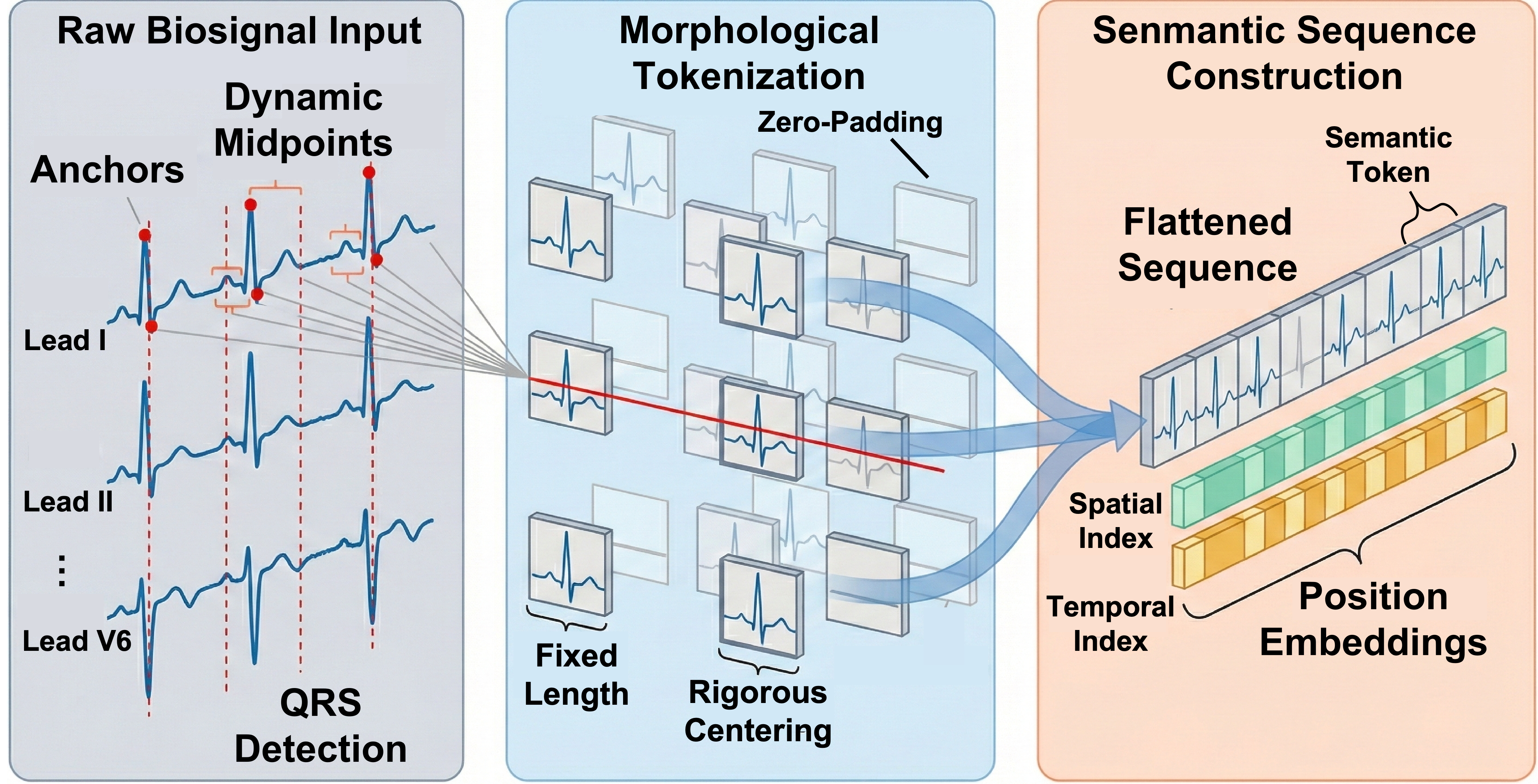}
\caption{Schematic illustration of the Biological Tokenization process. The continuous multi-lead ECG signal is first analyzed to identify R-peak timestamps which serve as temporal anchors. For each anchor, a fixed-length window is extracted and rigorously centered to align the QRS complex. These morphological units are then flattened into a sequence $\mathcal{H}$ and paired with explicit spatial lead indices and temporal position embeddings.}
\label{fig:tokenization}
\end{figure}

Shown in Fig~\ref{fig:tokenization}, the process begins by performing QRS detection on the primary lead to identify ventricular activation centroids which serve as global temporal anchors $\mathcal{A}$. For each anchor $\tau \in \mathcal{A}$ and each lead $c$, we extract a local waveform segment. To ensure consistent morphological alignment and suppress inter-beat baseline wander, we rigorously center the extracted physiological event within a zero-initialized array of fixed length $L$. This centering mechanism guarantees that the model's fundamental semantic unit preserves the physiological integrity of the P-QRS-T complex, regardless of heart rate variability.

The tokenizer then flattens these multi-channel segments into a unified sequence $\mathcal{H} \in \mathbb{R}^{S\times L}$ arranged hierarchically by lead index and chronological occurrence. To preserve spatiotemporal context, we concurrently generate auxiliary position identifiers including a spatial index vector recording the lead origin and a discrete temporal index vector derived from normalized R-peak timestamps.

\subsection{BEAT-Net Architecture}

\begin{figure}[!h]
\centering
\includegraphics[width=\linewidth]{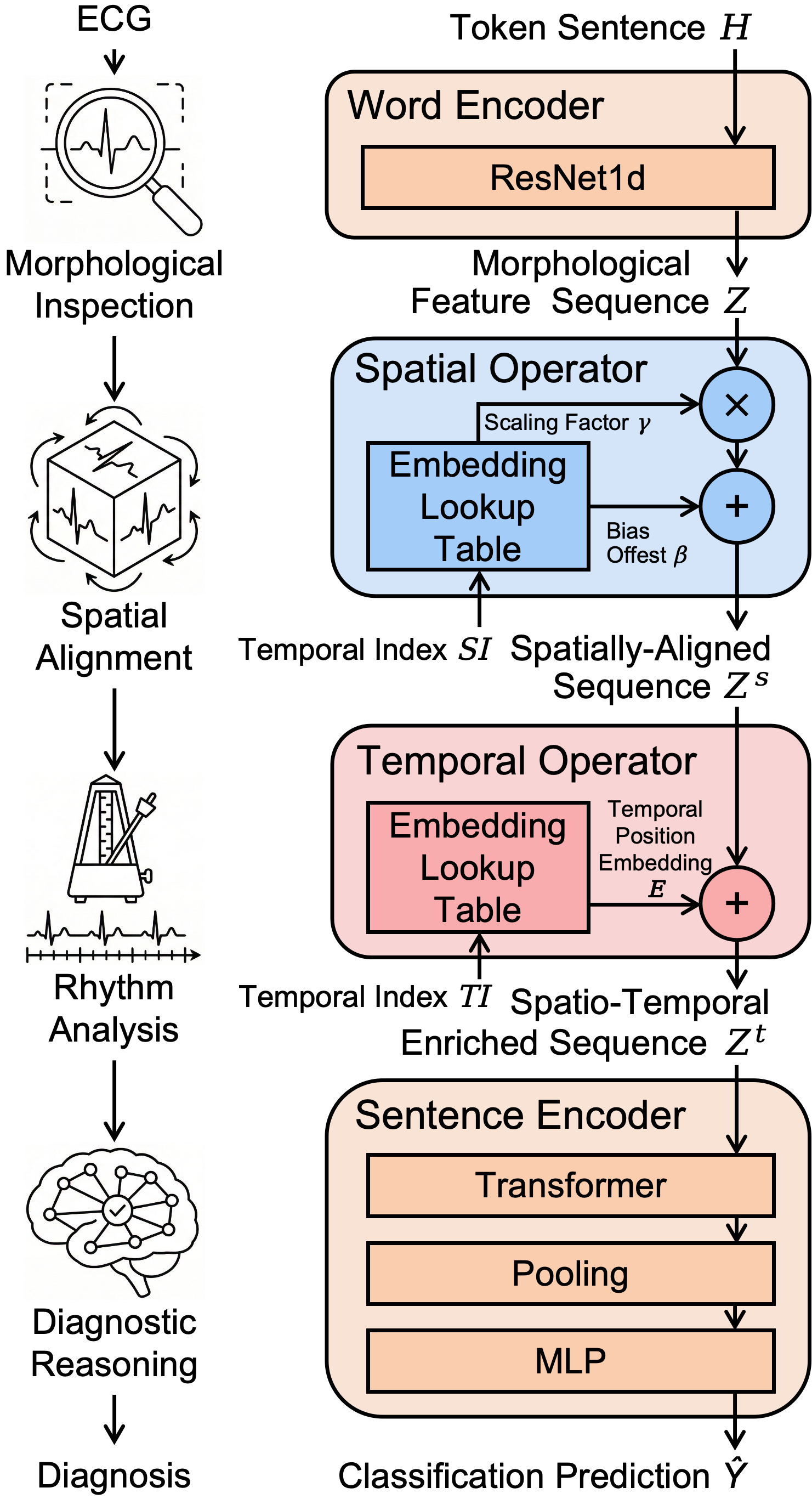}
\caption{Overview of the BEAT-Net framework. The Word Encoder first extracts morphological patterns into a Morphological Feature Sequence $Z$. Next, the Spatial Operator and Temporal Operator transform $Z$ into a Spatially-Aligned Sequence $Z^s$ and a Spatio-Temporal Enriched Sequence $Z^t$ using the Spatial Index $SI$ and Temporal Index $TI$, respectively. Finally, the Sentence Encoder utilizes a Transformer to capture long-range interactions from $Z^t$ for diagnostic prediction.}
\label{fig:beat-net}
\end{figure}

Shown in Fig~\ref{fig:beat-net} and formally defined in Algorithm~\ref{alg:beatnet}, the BEAT-Net framework processes the tokenized sequence through a multi-stage pipeline structured to emulate the cardiologist's workflow from local feature extraction to global decision-making.

\begin{algorithm}[h]
\caption{End-to-End Inference: Tokenization and BEAT-Net.}\label{alg:beatnet}
\begin{algorithmic}
\STATE 
\STATE {\textsc{PREPROCESS}}$(X)$ (Biological Tokenization)
\STATE \hspace{0.5cm}$ R \gets \textsc{QRSDetect}(X_{lead=0}) $
\STATE \hspace{0.5cm}$ \mathcal{H}, I_{lead}, I_{time} \gets \emptyset, \emptyset, \emptyset $
\STATE \hspace{0.5cm}\textbf{for} $ c = 1 \dots C $ \textbf{do}
\STATE \hspace{1.0cm}\textbf{for} $ r \in R $ \textbf{do}
\STATE \hspace{1.5cm}$ x_{token} \gets \textsc{Crop}(X_c, \text{center}=r, \text{len}=L) $
\STATE \hspace{1.5cm}$ \mathcal{H}.\text{append}(x_{token}) $
\STATE \hspace{1.5cm}$ I_{lead}.\text{append}(c); \ I_{time}.\text{append}(\text{Quantize}(r)) $
\STATE \hspace{0.5cm}$ H \gets \textsc{Stack}(\mathcal{H}) $
\STATE \hspace{0.5cm}$ M \gets (I_{lead} == 0) $
\STATE \hspace{0.5cm}\textbf{return} $ H, M, I_{lead}, I_{time} $
\STATE 
\STATE {\textsc{FORWARD}}$(H, M, I_{lead}, I_{time})$ (BEAT-Net Encoding) 
\STATE \hspace{0.5cm}$ Z \gets \mathcal{F}_{word}(H) $
\STATE \hspace{0.5cm}$ Z \gets Z \odot \gamma[I_{lead}] + \beta[I_{lead}] $
\STATE \hspace{0.5cm}$ Z \gets Z + \mathbf{E}_{time}[I_{time}] $
\STATE \hspace{0.5cm}$ Z_{seq} \gets \textsc{Transformer}(Z, \text{mask}=M) $
\STATE \hspace{0.5cm}$ z_{global} \gets \textsc{Sum}(Z_{seq} \cdot M) / (\textsc{Sum}(M) + \epsilon) $
\STATE \hspace{0.5cm}\textbf{return} $ \textsc{MLP}(z_{global}) $
\end{algorithmic}
\end{algorithm}

\subsubsection{Word Encoder}
The Word Encoder $\mathcal{F}_{word}$ acts as the initial feature extractor, mapping the raw tokenized sequence $H$ into a Morphological Feature Sequence $Z$ as
\begin{equation}
    z_i = \mathcal{F}{word}(h_i)
\end{equation}
where $\mathcal{F}{word}$ comprises a convolutional stem followed by stacked 1D residual blocks~\cite{he2016deep}. This stage corresponds to the morphological inspection phase, isolating intrinsic patterns such as QRS amplitude and width independently of lead configuration or heart rate.

\subsubsection{Spatial Operator}
Identical physiological events manifest with significant morphological variability across different leads depending on their anatomical orientation~\cite{de2021clinical}. To emulate the clinician's mental normalization of these view-dependent variations, we transform $Z$ into a Spatially-Aligned Sequence $Z^s$ using affine transformations as
\begin{equation}
    z^{s}i = z_i \odot \gamma_{SI} + \beta_{SI}
\end{equation}

where $\gamma_{SI}$ and $\beta_{SI}$ are learnable parameters corresponding to the lead specified by spatial index $SI$. This mechanism enforces a spatial inductive bias, enabling the model to learn a unified representation of cardiac activity invariant to viewing angles.

\subsubsection{Temporal Operator}
As emphasized in American Heart Association (AHA) guidelines~\cite{wagner2009aha}, precise ECG assessment requires analyzing not just morphology but the relative timing between cycles to detect arrhythmias. We inject this context into $Z^s$ to produce the Spatio-Temporal Enriched Sequence $Z^t$ as
\begin{equation}
    z^{t}i = z^{s}i + \mathbf{e}_{TI}
\end{equation}
where $\mathbf{e}_{TI}$ represents the learnable Temporal Position Embedding. This step effectively reconstructs the sequential structure of the heart rhythm, organizing the isolated tokens into a coherent chronological sequence.

\subsubsection{Sentence Encoder}
Finally, the Spatio-Temporal Enriched Sequence $Z^t$ is processed by a Transformer~\cite{vaswani2017attention} Encoder. This module serves as the global reasoning engine, analogous to a physician synthesizing all evidence to form a diagnosis. It resolves long-range dependencies across the entire recording to derive the final prediction $\hat{y}$ as
\begin{equation}
    \hat{y} = \mathcal{F}{\text{mlp}}( \text{Pool}( \mathcal{T}(Z^{t}) ) )
\end{equation}
where $\mathcal{T}$ denotes the Transformer layers and $\mathcal{F}{\text{mlp}}$ represents the classification head.

\section{Experiments}
This section details the data configurations and representative comparative models alongside specific implementation details to ensure robust and reproducible assessments.

\subsection{Datasets}
We assess model robustness and generalization using three large-scale 12-lead ECG benchmarks that cover diverse recording environments and pathological distributions.

\subsubsection{PTB-XL~\cite{wagner2020ptb}}
As the largest publicly available clinical dataset to date, PTB-XL comprises 21,837 ten-second records collected from 18,885 unique patients to ensure strict subject independence between splits. The dataset provides comprehensive multi-label annotations covering 71 distinct SCP-ECG statements which are categorically aggregated into 44 diagnostic, 19 form, and 12 rhythm classes. For benchmarking, we adhere to the standard evaluation protocol~\cite{strodthoff2020deep} by utilizing the official stratified assignment of folds 1--8 for training, fold 9 for validation, and fold 10 for testing, while focusing on diagnostic labels which can be hierarchically categorized into 5 superclasses and 23 subclasses according to different granularities.

\begin{figure}[ht]
\centering
\includegraphics[width=\linewidth]{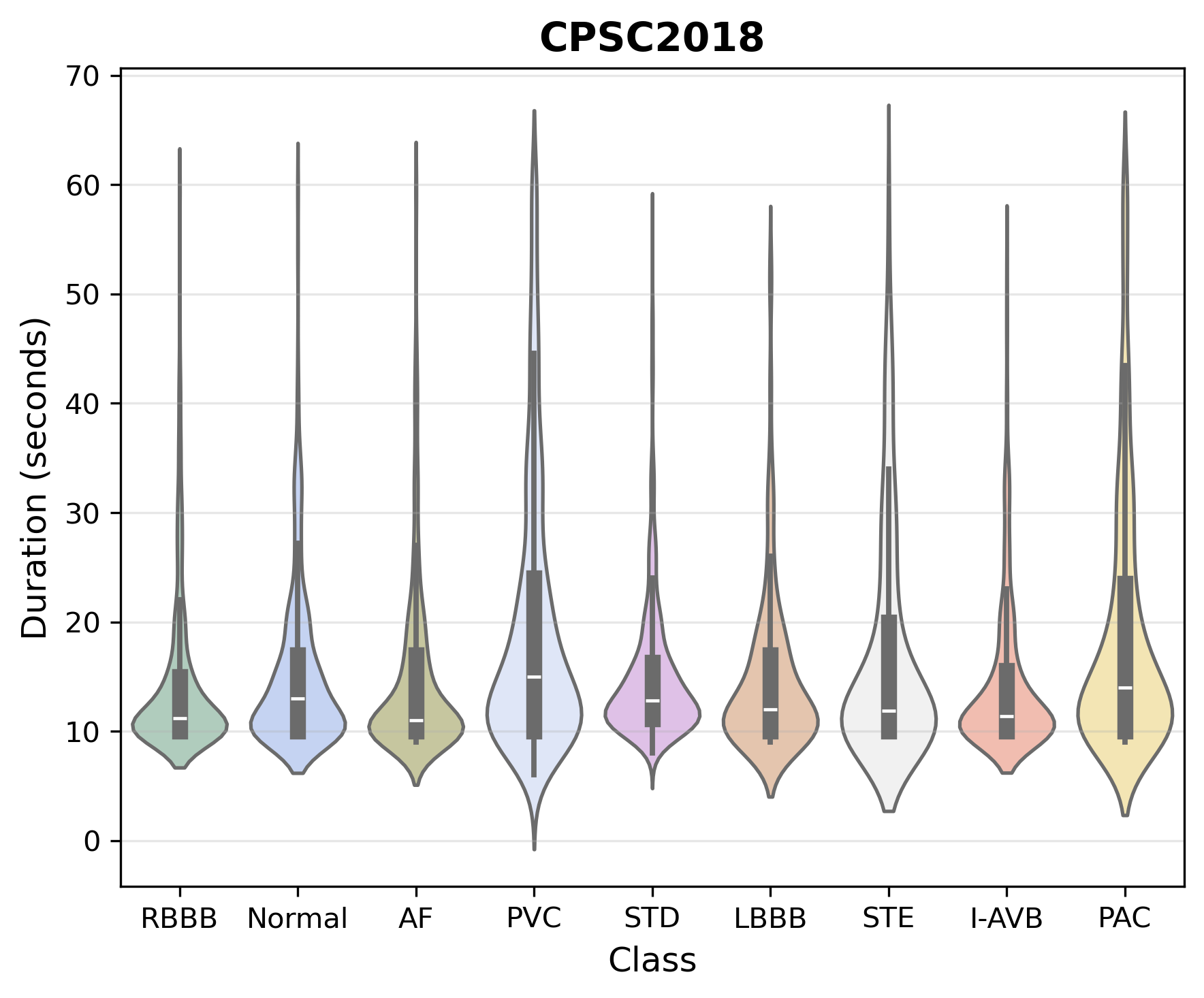}
\caption{Statistical characteristics of the CPSC2018 dataset. Violin plot illustrating the distribution of signal durations across different diagnostic classes in the CPSC2018 dataset. The width of each violin represents the density of samples at a given duration.}
\label{fig:CPSC2018}
\end{figure}

\subsubsection{CPSC2018~\cite{liu2018open}}
Originating from the 1st China Physiological Signal Challenge, this dataset contains 6,877 recordings collected from 11 distinct hospitals. Unlike standard fixed-length archives, CPSC2018 features variable signal durations ranging from 6 to 60 seconds, as illustrated in Fig \ref{fig:CPSC2018}, which presents significant challenges for batch processing. The annotations encompass nine categories including major rhythm disorders like Atrial Fibrillation (AF) and morphological abnormalities such as Bundle Branch Blocks (BBB). To handle the temporal variability, we standardize signal duration via truncation or padding, apply Min-Max normalization, and downsample to 100 Hz prior to a random 7:1:2 partition.

\subsubsection{CSN~\cite{zheng2020optimal,zheng2022large}}
To evaluate performance on large-scale data, we utilize the Chapman-Shaoxing-Ningbo (CSN) database which includes approximately 45,000 records sampled at a high fidelity of 500 Hz. This cohort undergoes rigorous expert validation to label complex arrhythmias including Sinus Bradycardia (SB) and Supraventricular Tachycardia (ST). Given the high sampling rate and potential acquisition noise, we downsample the signals to 100 Hz and apply a 5th-order Butterworth bandpass filter between 0.67 and 40 Hz to suppress baseline wander and high-frequency artifacts before splitting the data according to a 7:1:2 ratio.

\subsection{Baselines}
We benchmark BEAT-Net against a diverse set of competitive methods, ranging from canonical 1D-CNN architectures established in the comprehensive ECG benchmarking study~\cite{strodthoff2020deep} to the foundation model pre-trained on large-scale datasets.

\subsubsection{xresnet1d101~\cite{he2019bag}}
As a representative of deep residual learning, this model adapts the 101-layer ResNet architecture by incorporating the structural refinements proposed in the Bag of Tricks study. Unlike standard implementations, xresnet1d101 modifies the convolutional stem and downsampling blocks to replace stride-2 convolutions with blur pooling operations. These architectural modifications preserve information flow during downsampling and allow the network to scale to significant depths without suffering from performance degradation or training instability.

\subsubsection{resnet1d\_wang~\cite{wang2017time}}
We also evaluate the specific ResNet variant proposed by Wang et al. which serves as a robust baseline for time series classification. This architecture utilizes a Fully Convolutional Network backbone augmented with residual shortcuts to facilitate gradient propagation across layers. To enhance the receptive field, the model incorporates large initial kernels. Furthermore, it replaces standard pooling with a concatenation strategy that fuses global average and global maximum statistics, thereby capturing both dominant and background signal features.

\subsubsection{inception1d~\cite{ismail2020inceptiontime}}
To address the multi-scale nature of cardiac electrophysiology, we utilize the InceptionTime architecture. This model comprises an ensemble of Inception modules that execute parallel convolutions with varying kernel lengths. This design enables the simultaneous extraction of high-frequency morphological details and low-frequency rhythmic patterns. Furthermore, the architecture integrates bottleneck layers to reduce dimensionality and residual connections to ensure convergence. Collectively, these features establish a strong baseline for capturing complex temporal dependencies without the need for recurrent units.

\subsubsection{HeartLang~\cite{jin2025reading}}
To benchmark against the emerging paradigm of physiological foundation models, we evaluate HeartLang, a state-of-the-art self-supervised framework. Diverging from standard continuous time-series modeling, HeartLang employs a QRS-tokenizer to discretize signals into heartbeat-centric units. These tokens are subsequently processed by the ST-ECGFormer, a Transformer backbone enriched with learnable spatio-temporal embeddings to resolve complex cardiac dependencies. In our experiments, we utilize the official model weights pre-trained via masked sentence modeling and vector-quantized reconstruction. We deploy the backbone as a fixed feature extractor, where the resulting high-dimensional semantic representations are fed into a Multi-Layer Perceptron (MLP) for classification. This baseline serves to evaluate the transferability of large-scale self-supervised priors compared to our supervised biomimetic approach.

\begin{table*}[!ht] 
\centering 
\caption{Quantitative performance and model complexity comparison of BEAT-Net against leading 1D-CNN architectures and foundation model baseline across the PTB-XL, CPSC2018, and CSN datasets. The performance evaluation metric is AUC. The best results for each category are marked in \textbf{bold}. Note that the parameter counts (in millions, M) are reported based on the model configuration for the PTB-XL All task.} 
\label{tab:main_results} 

\begin{tabular*}{\linewidth}{l@{\extracolsep{\fill}}c cccccc cc}
    \toprule
    \multirow{2}{*}{\textbf{Models}} & \multirow{2}{*}{\textbf{Parameters (M)}} & \multicolumn{6}{c}{\textbf{PTB-XL}} & \multirow{2}{*}{\textbf{CPSC2018}} & \multirow{2}{*}{\textbf{CSN}} \\
    \cmidrule(lr){3-8}
     & & \textbf{All} & \textbf{Diag} & \textbf{Sup} & \textbf{Sub} & \textbf{Form} & \textbf{Rhythm} & & \\
    \midrule
    xresnet1d101 & 2.07 & \textbf{0.925} & \textbf{0.937} & 0.928 & \textbf{0.939} & 0.896 & 0.957 & 0.919 & 0.927 \\
    resnet1d\_wang & 1.37 & 0.919 & 0.936 & \textbf{0.930} & 0.928 & 0.880 & 0.946 & 0.941 & 0.935 \\ 
    inception1d & 0.71 & 0.925 & 0.931 & 0.921 & 0.930 & 0.899 & 0.953 & 0.937 & 0.929 \\
    HeartLang+MLP & 39.49 & 0.904 & 0.888 & 0.883 & 0.870 & 0.832 & \textbf{0.965} & 0.849 & 0.823 \\
    \midrule
    \textbf{BEAT-Net} & \textbf{0.70} & 0.924 & 0.936 & \textbf{0.931} & 0.937 & \textbf{0.901} & 0.952 & \textbf{0.949} & \textbf{0.942} \\
    \bottomrule
\end{tabular*}
\end{table*}

\subsection{Implementation}
We implemented all models using the PyTorch framework on a workstation equipped with a single NVIDIA RTX A6000 GPU. Parameter optimization utilized the AdamW algorithm~\cite{loshchilov2017decoupled} with an initial learning rate of $1\times10^{-3}$ and a batch size of 128. To mitigate overfitting and ensure optimal convergence, we incorporated an early stopping strategy that halts training when validation loss fails to improve after a defined patience period. We minimized the Binary Cross-Entropy with Logits Loss for all multi-label classification tasks while employing the macro-averaged Area Under the Curve (AUC) as the primary metric for performance monitoring and best-model checkpointing.

\section{Results} 
To rigorously evaluate the diagnostic performance and generalization capability of the proposed framework, we established a comprehensive experimental protocol using diverse large-scale benchmarks. The assessment begins by benchmarking against representative 1D-CNN architectures and a state-of-the-art foundation model baseline, confirming the necessity of each architectural component through ablation studies. We then demonstrate the model's exceptional data efficiency in low-resource regimes and conclude by visualizing the physiological interpretability of the learned attention mechanisms.

\subsection{Baselines Comparison}

We benchmark BEAT-Net against representative 1D-CNN architectures and the HeartLang foundation model, with quantitative results detailed in Table~\ref{tab:main_results}. The empirical evidence highlights not only the diagnostic accuracy but also the superior parameter efficiency of our biomimetic architecture. Remarkably, BEAT-Net achieves performance parity with the heavyweight xresnet1d101, with AUC scores of 0.924 versus 0.925, while utilizing only roughly one-third of the parameters at 0.7M versus 2.06M. Even when juxtaposed with the efficient Inception1D model, our framework maintains a lighter computational footprint while delivering competitive scores.

Comparison with Foundation Model. The inclusion of the HeartLang baseline provides a critical perspective on the trade-off between scale and structural priors. HeartLang achieves the highest Rhythm score of 0.965. It is crucial to note that, like BEAT-Net, HeartLang also employs a QRS-centered tokenization strategy that treats heartbeats as semantic tokens~\cite{jin2025reading}. This shared success on the Rhythm task strongly validates our hypothesis that treating the heartbeat as a fundamental semantic unit rather than using fixed-window slicing is key to capturing the long-range dependencies required for rhythm analysis.

However, despite this shared tokenizer, the frozen-feature approach of the foundation model significantly underperforms in Morphological tasks, specifically Form, achieving an AUC of only 0.832 compared to BEAT-Net's 0.901. This 6.9\% gap suggests that without task-specific supervision, the pre-trained features lose fine-grained waveform details essential for identifying local pathologies. Furthermore, HeartLang requires over 39 million parameters, approximately 56 times larger than BEAT-Net. On the CPSC2018 and CSN datasets, BEAT-Net demonstrates superior generalization, outperforming HeartLang by over 10\%, proving that our lightweight, physiologically grounded architecture offers a more efficient and robust alternative to massive foundation models for clinical deployment.

\subsection{Ablation Study}
To validate the effectiveness of the proposed framework, we conducted a systematic ablation study covering the input tokenization strategy and the specialized architectural modules. The results substantiate that BEAT-Net is not merely a generic Transformer but a specialized framework where biological alignment and the hierarchical application of physiological operators are indispensable.

\begin{figure}[ht]
    \centering
    \includegraphics[width=\linewidth]{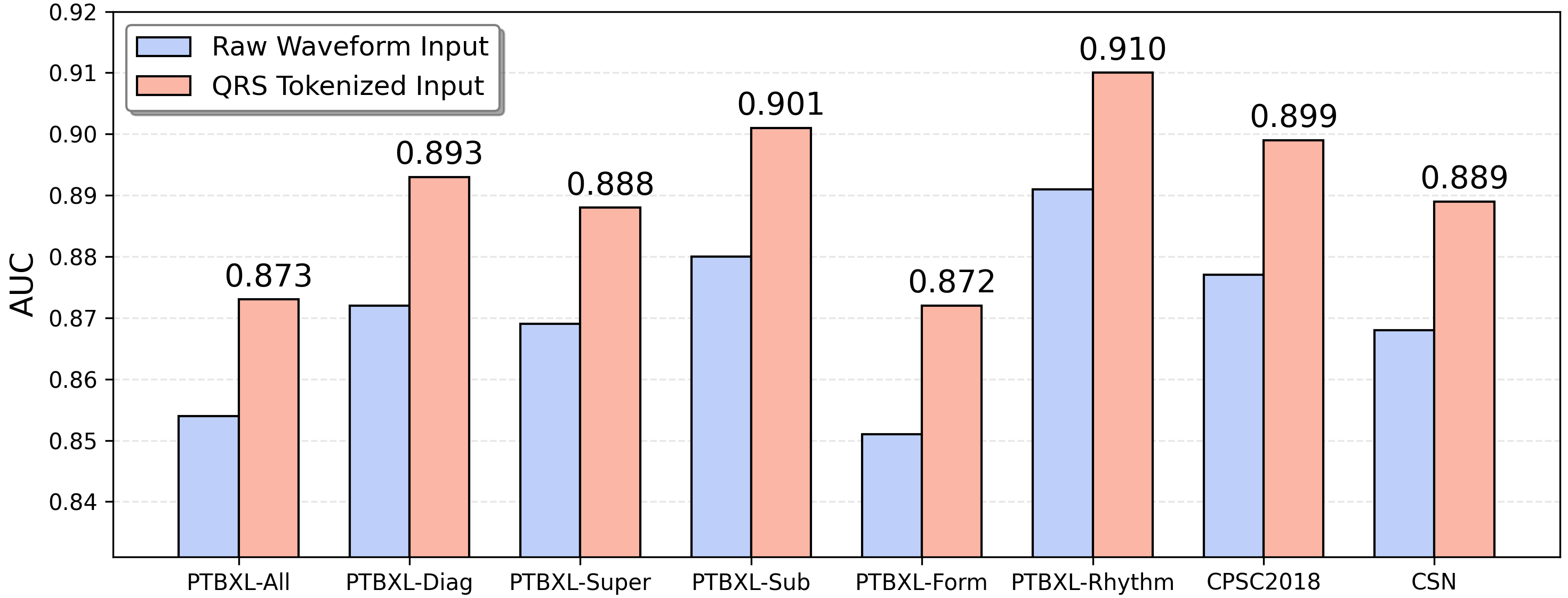}
    \caption{Impact of input tokenization strategies. The bar chart compares the AUC performance of the proposed QRS-centered tokenization against the raw waveform patching baseline across all PTB-XL tasks.}
    \label{fig:tokenization_effect}
\end{figure}

\subsubsection{Efficacy of Biological Tokenization}
We first evaluate the foundational impact of our input strategy in Fig~\ref{fig:tokenization_effect}. We compare our QRS-centered tokenization against standard sliding-window patching methods commonly used in vision-based approaches. The quantitative superiority of our strategy confirms that biological alignment is critical for data efficiency. By rigorously centering the QRS complex, we ensure that each token encapsulates a semantically complete heartbeat unit. In contrast, rigid waveform patching often arbitrarily severs physiological structures like P-waves or T-waves across token boundaries which forces the model to waste capacity relearning these fragmented patterns.

\subsubsection{Impact of Spatio-Temporal Operators}
Complementing the input strategy, we validate the architectural design by systematically removing key components as detailed in Table~\ref{tab:ablation_study}. Reducing BEAT-Net to a standard multi-level Transformer by excising both Spatial and Temporal Operators causes performance to collapse to an AUC of 0.873 on PTB-XL All tasks. This 5.1\% degradation indicates that pure self-attention mechanisms lack the inductive bias necessary to navigate the complex topology of 12-lead ECGs without explicit guidance.

\begin{table}[!ht]
    \centering
    \caption{Ablation study of BEAT-Net components across the PTB-XL, CPSC2018, and CSN datasets. The performance evaluation metric is AUC. The best results are \textbf{bolded} (w/o: without; ST Op.: Spatio-Temporal Operator).}
    \label{tab:ablation_study}
    
    \fontsize{8}{9}\selectfont
    \setlength{\tabcolsep}{0pt}
    
    \begin{tabular*}{\linewidth}{l@{\extracolsep{\fill}}cccccccc}
        \toprule
        \multirow{2}{*}{\textbf{Models}} & \multicolumn{6}{c}{\textbf{PTB-XL}} & \multirow{2}{*}{\textbf{CPSC2018}} & \multirow{2}{*}{\textbf{CSN}} \\
        \cmidrule(lr){2-7}
        & \textbf{All} & \textbf{Diag} & \textbf{Sup} & \textbf{Sub} & \textbf{Form} & \textbf{Rhythm} \\
        \midrule
        
        w/o ST Op. & 0.873 & 0.893 & 0.888 & 0.901 & 0.872 & 0.910 & 0.908 & 0.900 \\
        
        w/o S Op. & 0.899 & 0.914 & 0.911 & 0.917 & 0.881 & 0.931 & 0.927 & 0.919 \\
        
        w/o T Op. & 0.901 & 0.916 & 0.907 & 0.916 & 0.879 & 0.933 & 0.928 & 0.921 \\
        
        \textbf{BEAT-Net} & \textbf{0.924} & \textbf{0.936} & \textbf{0.931} & \textbf{0.937} & \textbf{0.901} & \textbf{0.952} & \textbf{0.949} & \textbf{0.942} \\
        
        \bottomrule
    \end{tabular*}
\end{table}

\begin{figure*}[!ht]
    \centering
    \includegraphics[width=\linewidth]{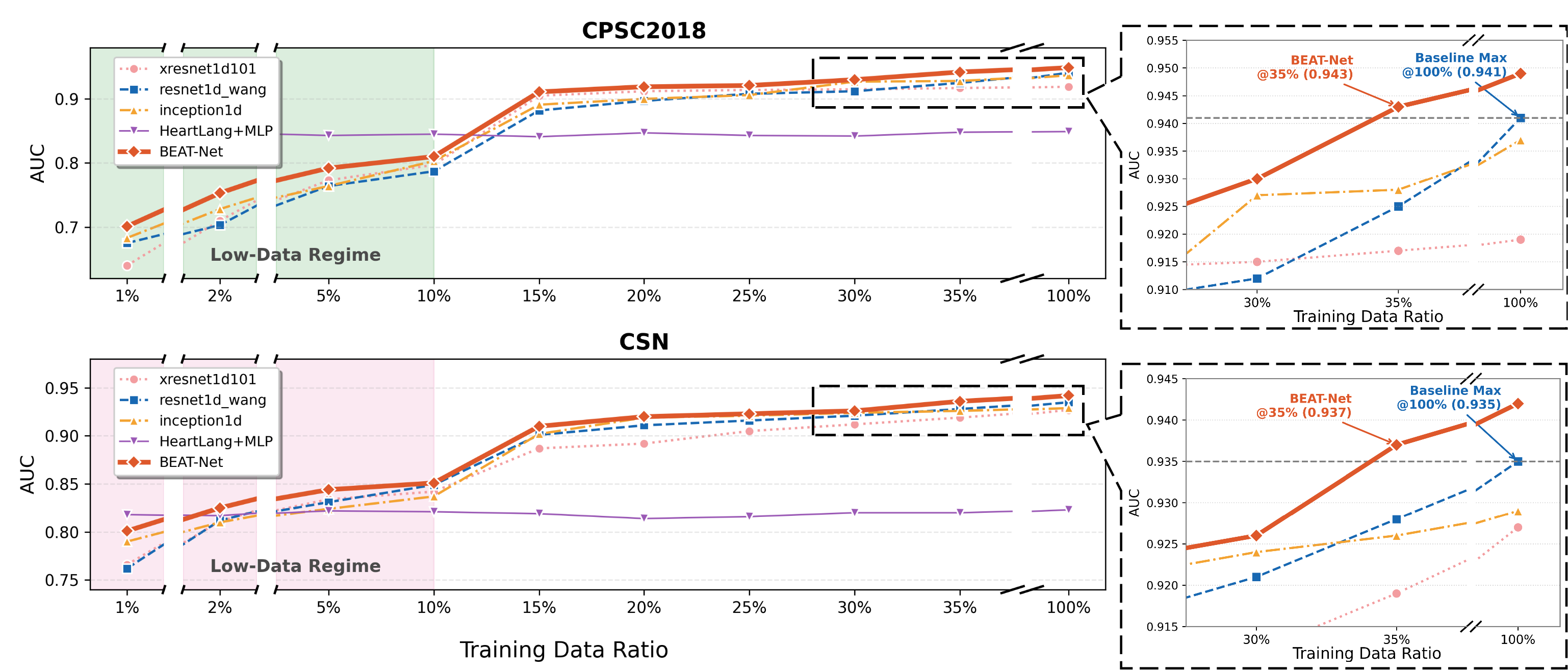}
    \caption{Data efficiency analysis on the CPSC2018 and CSN datasets. The main plots illustrate model performance across the full training spectrum. The shaded regions highlight the low-data regime 1\%--10\% where BEAT-Net demonstrates rapid learning capability. Notably, the HeartLang+MLP baseline exhibits a characteristically flat performance curve, indicating a heavy reliance on frozen pre-trained features that fail to adapt with increased supervision. In contrast, BEAT-Net surpasses this foundation model baseline with only a fraction of data and achieves parity with fully-supervised CNNs using just 35\% of the training samples.}
    \label{fig:data_efficiency}
\end{figure*}

\begin{figure}[!ht]
    \centering
    \includegraphics[width=\linewidth]{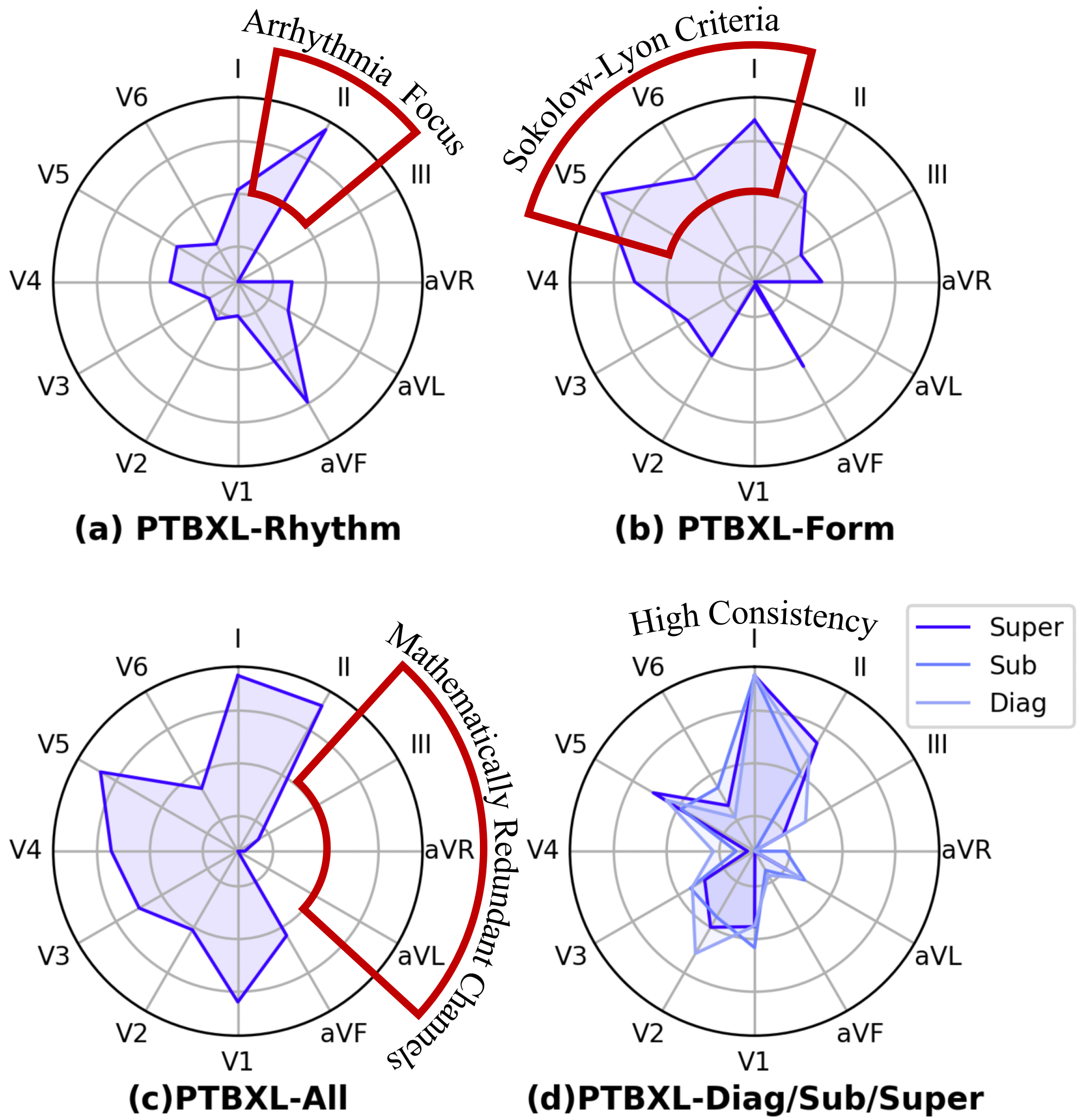}
    \caption{Visualization of learned spatial attention patterns. The radar charts depict the aggregate attention weights assigned to each of the 12 leads on the PTB-XL dataset. (a) Lead II exhibits distinct dominance in Rhythm classification tasks. (b) Leads I, V5, and V6 are primarily emphasized in Form classification tasks. (c) The All task category displays a physiologically efficient profile where the model integrates critical views from Leads II, I, V5, and V6 while suppressing the mathematically redundant channels of Leads III, aVR, and aVL. (d) The overall attention profiles remain consistent across the varying granularities of Superclass, Subclass, and Diagnostic hierarchies.}
    \label{fig:radar_chart}
\end{figure}

Single-component ablations further clarify the distinct roles of each operator. Removing the Spatial Operator drops the Form AUC to 0.881 which suggests that without lead-specific normalization, the model struggles to reconcile morphological variations across different spatial views. Conversely, excluding the Temporal Operator specifically impairs sequential tracking capabilities. This deficiency is evidenced by the sharp decline in Rhythm AUC from 0.952 to 0.933. Such degradation occurs because the model loses the explicit positional context required to identify arrhythmias defined by irregular beat intervals, such as AF.

\subsection{Data Efficiency Analysis}

The acquisition of annotated datasets remains a primary bottleneck in clinical AI~\cite{rajpurkar2022ai}. We evaluated the robustness of BEAT-Net in low-resource settings by training on subsets of the CPSC2018 and CSN datasets ranging from 1\% to 100\%, benchmarking against both CNNs and the HeartLang foundation model.

\textbf{Comparison with Foundation Model.} As illustrated in Fig~\ref{fig:data_efficiency}, the HeartLang+MLP baseline remains nearly flat on both CPSC2018 and CSN, staying around 0.84 AUC on CPSC2018 and around 0.82 AUC on CSN across sample sizes. This phenomenon is characteristic of frozen backbone strategies. While massive pre-training provides a strong initial feature space that yields decent zero-shot or few-shot performance from 1\% data, the fixed representations lack the plasticity to adapt to domain-specific nuances even when more annotated data becomes available.

BEAT-Net's Rapid Convergence. In stark contrast, BEAT-Net trained entirely from scratch demonstrates a steep and efficient learning curve. On CSN, BEAT-Net surpasses the foundation model with only 5\%--10\% of the training data, while on CPSC2018 it does so from 15\% onward. This rapid convergence suggests that our biomimetic architectural priors, including tokenization and modular operators, provide a more effective inductive bias for clinical diagnosis than generic pre-trained features. By structurally enforcing a diagnostic reasoning process that mirrors clinical workflows, BEAT-Net bypasses the data-intensive stage of learning basic waveform morphology and focuses immediately on pathological discrimination.

\textbf{Efficiency against CNNs.} Furthermore, BEAT-Net achieves performance parity with fully supervised CNN baselines using significantly fewer samples. On the CSN dataset, the model attains an AUC of 0.936 using only 35\% of training samples, outperforming the strongest baseline trained on the complete dataset. This trend holds true on CPSC2018, where 35\% of the data yields an AUC of 0.942, exceeding the fully trained resnet1d\_wang baseline. These results indicate that BEAT-Net effectively lowers annotation requirements by approximately 65\% while delivering superior diagnostic accuracy compared to both data-hungry CNNs and rigid foundation models.

\subsection{Interpretability and Clinical Alignment}

The black-box nature~\cite{rudin2019stop} of deep neural networks often hinders their adoption in clinical settings where transparency is paramount. BEAT-Net alleviates this issue through the Spatial Operator's transparent attention mechanism which offers a window into the model's decision-making process. As visualized in Fig~\ref{fig:radar_chart}, the model autonomously acquires attention patterns that align with established cardiological heuristics without explicit supervision.

In the rhythm analysis presented in Fig \ref{fig:radar_chart}(a), the model exhibits a pronounced bias toward Lead II. This effectively mirrors the clinical standard where Lead II is preferred for identifying P-waves and diagnosing arrhythmias due to its alignment with the primary vector of atrial depolarization~\cite{sandau2017update}. Conversely, the morphological tasks in Fig~\ref{fig:radar_chart}(b) trigger a distinct shift toward leads I, V5, and V6. This pattern corresponds directly with diagnostic indices such as the Sokolow-Lyon criteria $SV1 + RV5/V6$ used to assess Left Ventricular Hypertrophy (LVH) and BBB~\cite{sokolow1949ventricular}. Such alignment confirms that the Spatial Operator does not merely memorize statistical correlations but correctly identifies the anatomical perspectives most relevant to structural anomalies.

Crucially, the aggregated view in Fig~\ref{fig:radar_chart}(c) reveals a physiologically efficient rather than uniform distribution, providing compelling evidence of the model's ability to minimize information redundancy. We observe that Lead III, aVR, and aVL receive notably lower attention weights compared to Leads I, II, and V1--V6. This sparsity adheres to the intrinsic physics of the 12-lead system. According to Einthoven's Law, Lead III is mathematically derivable from Leads I and II through the equation $III = II - I$~\cite{kligfield2007recommendations}. This linear dependency renders it computationally redundant given the model's strong focus on the primary leads. Similarly, the attenuation of aVR weights aligns with its definition as an augmented lead derived from the inverse average of Leads I and II via the expression $aVR = -(I + II)/2$~\cite{goldberger2023goldberger, yang2018independent}. Finally, the reduced focus on aVL can be attributed to its spatial redundancy as a contiguous high-lateral lead sharing a nearly identical vector orientation with Lead I~\cite{wagner2009aha}. By pruning these linearly dependent or spatially overlapping channels, BEAT-Net demonstrates a high-level abstraction capability by effectively isolating the independent basis vectors required for complete spatiotemporal reconstruction.

In Fig~\ref{fig:radar_chart}(d), the attention profiles remain highly consistent across the Superclass, Subclass, and Diagnostic hierarchies. This visual similarity suggests that the learned features are fundamental representations of cardiac physiology rather than task-specific artifacts. By spontaneously reproducing human expert attention patterns, BEAT-Net provides interpretable evidence that validates its diagnostic logic and fosters clinical trust.

\section{Discussion}

BEAT-Net demonstrates that explicit physiological structure provides a stronger inductive bias than either signal-agnostic CNNs or large-scale pre-training. By achieving performance parity with xresnet1d101 across 0.924 versus 0.925 AUC while using 95 percent fewer parameters, we show that model capacity is not the limiting factor in ECG diagnosis. Rather, the quality of inductive bias determines efficiency. The QRS-centered tokenization strategy preserves physiological coherence by treating the P-QRS-T complex as the fundamental semantic unit, enabling the model to focus learning on pathological discrimination rather than wasting capacity on reconstructing fragmented waveforms. Ablation studies confirm this design is critical, as arbitrary patch slicing produces substantial performance degradation, validating our biomimetic approach.

The data efficiency curves reveal a fundamental distinction between models with architectural priors and frozen pre-trained features. HeartLang+MLP exhibits a flat learning curve, constrained by its fixed representations, while BEAT-Net improves steadily with increasing supervision. BEAT-Net surpasses HeartLang at 5-10 percent on CSN and from 15 percent onward on CPSC2018, reflecting differences in signal quality across datasets. More importantly, BEAT-Net reaches full supervised performance using only 35 percent of training data on both datasets, simultaneously achieving superior accuracy over CNNs trained on complete data, a rare triple efficiency in parameters, data, and computation.

Learned attention patterns spontaneously align with established clinical heuristics without explicit supervision. Lead II dominance for rhythm tasks mirrors clinical preference for P-wave identification, while lateral-lead emphasis for morphology reflects diagnostic criteria like Sokolow-Lyon indices. The model downweights mathematically redundant leads that are linearly dependent via Einthoven's and Goldberger's laws, demonstrating internalization of physiological principles. These emergent behaviors validate that BEAT-Net learns genuine cardiac physiology rather than spurious correlations, supporting clinical adoption.

Limitations exist in that the pipeline depends on reliable QRS detection, which can fail in severe arrhythmias or high-noise recordings. Integration of differentiable tokenization or joint detection-diagnosis modules could address this. Extending evaluation to ambulatory recordings with realistic artifacts remains essential for deployment validation. Future work should also investigate whether learned lead-weighting patterns could inform accelerated ECG acquisition protocols for point-of-care settings.

\section{Conclusion}

This work establishes that domain-aligned architectural design achieves superior efficiency and interpretability compared to scale-based approaches in specialized medical domains. BEAT-Net demonstrates this principle by achieving performance parity with CNN baselines while reducing parameters by 95 percent, exceeding a 39.5-million-parameter foundation model despite being 56 times smaller, and attaining full supervised performance with only 35 percent of annotated data. Critically, learned attention mechanisms spontaneously align with established clinical heuristics without supervision, indicating that BEAT-Net captures genuine physiological principles rather than spurious correlations. These results establish a new benchmark for parameter and data efficiency in medical AI while maintaining the interpretability essential for clinical deployment, opening a research direction toward efficient, transparent clinical AI systems grounded in domain expertise.

\section*{References}

\bibliographystyle{IEEEtran}
\bibliography{ref}

\begin{thebibliography}{10}
\providecommand{\url}[1]{#1}
\csname url@samestyle\endcsname
\providecommand{\newblock}{\relax}
\providecommand{\bibinfo}[2]{#2}
\providecommand{\BIBentrySTDinterwordspacing}{\spaceskip=0pt\relax}
\providecommand{\BIBentryALTinterwordstretchfactor}{4}
\providecommand{\BIBentryALTinterwordspacing}{\spaceskip=\fontdimen2\font plus
\BIBentryALTinterwordstretchfactor\fontdimen3\font minus
  \fontdimen4\font\relax}
\providecommand{\BIBforeignlanguage}[2]{{%
\expandafter\ifx\csname l@#1\endcsname\relax
\typeout{** WARNING: IEEEtran.bst: No hyphenation pattern has been}%
\typeout{** loaded for the language `#1'. Using the pattern for}%
\typeout{** the default language instead.}%
\else
\language=\csname l@#1\endcsname
\fi
#2}}
\providecommand{\BIBdecl}{\relax}
\BIBdecl

\bibitem{tsao2023heart}
C.~W. Tsao, A.~W. Aday, Z.~I. Almarzooq \emph{et~al.}, ``{Heart} disease and
  stroke statistics—2023 update: a report from the {American Heart
  Association},'' \emph{Circulation}, vol. 147, no.~8, pp. e93--e621, 2023.

\bibitem{writing20212021}
W.~C. Members, M.~Gulati, P.~D. Levy \emph{et~al.}, ``2021
  {AHA/ACC/ASE/CHEST/SAEM/SCCT/SCMR} guideline for the evaluation and diagnosis
  of chest pain,'' \emph{J. Am. Coll. Cardiol.}, vol.~78, no.~22, pp.
  e187--e285, 2021.

\bibitem{dosovitskiy2020image}
A.~Dosovitskiy, ``An image is worth 16x16 words: Transformers for image
  recognition at scale,'' \emph{arXiv preprint arXiv:2010.11929}, 2020.

\bibitem{ansari2025survey}
M.~Y. Ansari, M.~Yaqoob, M.~Ishaq \emph{et~al.}, ``A survey of transformers and
  large language models for {ECG} diagnosis: advances, challenges, and future
  directions,'' \emph{Artif. Intell. Rev.}, vol.~58, no.~9, p. 261, 2025.

\bibitem{hong2019mina}
S.~Hong, C.~Xiao, T.~Ma, H.~Li, and J.~Sun, ``{MINA}: multilevel
  knowledge-guided attention for modeling electrocardiography signals,''
  \emph{arXiv preprint arXiv:1905.11333}, 2019.

\bibitem{lan2022intra}
X.~Lan, D.~Ng, S.~Hong, and M.~Feng, ``Intra-inter subject self-supervised
  learning for multivariate cardiac signals,'' in \emph{Proc. AAAI Conf. Artif.
  Intell.}, vol.~36, no.~4, 2022, pp. 4532--4540.

\bibitem{rudin2019stop}
C.~Rudin, ``Stop explaining black box machine learning models for high stakes
  decisions and use interpretable models instead,'' \emph{Nat. Mach. Intell.},
  vol.~1, no.~5, pp. 206--215, 2019.

\bibitem{weng2025self}
W.~Weng, Y.~Gu, S.~Guo, Y.~Ma, Z.~Yang, Y.~Liu, and Y.~Chen, ``Self-supervised
  learning for electroencephalogram: A systematic survey,'' \emph{ACM Comput.
  Surv.}, vol.~57, no.~12, pp. 1--38, 2025.

\bibitem{jin2025reading}
J.~Jin, H.~Wang, H.~Li, J.~Li, J.~Pan, and S.~Hong, ``Reading your heart:
  Learning {ECG} words and sentences via pre-training {ECG} language model,''
  \emph{arXiv preprint arXiv:2502.10707}, 2025.

\bibitem{goldberger2018goldberger}
A.~Goldberger, \emph{{Goldberger}'s clinical electrocardiography}.\hskip 1em
  plus 0.5em minus 0.4em\relax Elsevier, 2018, vol.~10.

\bibitem{he2016deep}
K.~He, X.~Zhang, S.~Ren, and J.~Sun, ``Deep residual learning for image
  recognition,'' in \emph{Proc. IEEE Conf. Comput. Vis. Pattern Recognit.
  (CVPR)}, 2016, pp. 770--778.

\bibitem{vaswani2017attention}
A.~Vaswani, N.~Shazeer, N.~Parmar \emph{et~al.}, ``Attention is all you need,''
  \emph{Adv. Neural Inf. Process. Syst.}, vol.~30, 2017.

\bibitem{wagner2020ptb}
P.~Wagner, N.~Strodthoff, R.-D. Bousseljot \emph{et~al.}, ``{PTB-XL}, a large
  publicly available electrocardiography dataset,'' \emph{Sci. Data}, vol.~7,
  no.~1, pp. 1--15, 2020.

\bibitem{liu2018open}
F.~Liu, C.~Liu, L.~Zhao, X.~Zhang \emph{et~al.}, ``An open access database for
  evaluating the algorithms of electrocardiogram rhythm and morphology
  abnormality detection,'' \emph{J. Med. Imag. Health Informat.}, vol.~8,
  no.~7, pp. 1368--1373, 2018.

\bibitem{zheng2020optimal}
J.~Zheng, H.~Chu, D.~Struppa \emph{et~al.}, ``Optimal multi-stage arrhythmia
  classification approach,'' \emph{Sci. Rep.}, vol.~10, no.~1, p. 2898, 2020.

\bibitem{zheng2022large}
J.~Zheng, H.~Guo, and H.~Chu, ``A large scale 12-lead electrocardiogram
  database for arrhythmia study (version 1.0.0),'' \emph{PhysioNet}, vol.~23,
  p.~7, 2022, [Online]. Available:
  https://physionet.org/content/ecg-arrhythmia/1.0.0/.

\bibitem{strodthoff2020deep}
N.~Strodthoff, P.~Wagner, T.~Schaeffter, and W.~Samek, ``Deep learning for
  {ECG} analysis: Benchmarks and insights from {PTB-XL},'' \emph{IEEE J.
  Biomed. Health Informat.}, vol.~25, no.~5, pp. 1519--1528, 2020.

\bibitem{sandau2017update}
K.~E. Sandau, M.~Funk, A.~Auerbach \emph{et~al.}, ``Update to practice
  standards for electrocardiographic monitoring in hospital settings: a
  scientific statement from the {American Heart Association},''
  \emph{Circulation}, vol. 136, no.~19, pp. e273--e344, 2017.

\bibitem{sokolow1949ventricular}
M.~Sokolow and T.~P. Lyon, ``The ventricular complex in left ventricular
  hypertrophy as obtained by unipolar precordial and limb leads,'' \emph{Amer.
  Heart J.}, vol.~37, no.~2, pp. 161--186, 1949.

\bibitem{jaya2025advances}
A.~Jaya~Prakash, A.~Nasreddine~Belkacem \emph{et~al.}, ``Advances in machine
  and deep learning for {ECG} beat classification: a systematic review,''
  \emph{Front. Digit. Health}, vol.~7, p. 1649923, 2025.

\bibitem{petmezas2022state}
G.~Petmezas, L.~Stefanopoulos, V.~Kilintzis \emph{et~al.}, ``State-of-the-art
  deep learning methods on electrocardiogram data: systematic review,''
  \emph{JMIR Med. Informat.}, vol.~10, no.~8, p. e38454, 2022.

\bibitem{kiranyaz2015real}
S.~Kiranyaz, T.~Ince, and M.~Gabbouj, ``Real-time patient-specific {ECG}
  classification by 1-{D} convolutional neural networks,'' \emph{IEEE Trans.
  Biomed. Eng.}, vol.~63, no.~3, pp. 664--675, 2015.

\bibitem{hannun2019cardiologist}
A.~Y. Hannun, P.~Rajpurkar, M.~Haghpanahi \emph{et~al.}, ``Cardiologist-level
  arrhythmia detection and classification in ambulatory electrocardiograms
  using a deep neural network,'' \emph{Nat. Med.}, vol.~25, no.~1, pp. 65--69,
  2019.

\bibitem{kiranyaz2015convolutional}
S.~Kiranyaz, T.~Ince, R.~Hamila, and M.~Gabbouj, ``Convolutional neural
  networks for patient-specific {ECG} classification,'' in \emph{Proc. 37th
  Annu. Int. Conf. IEEE Eng. Med. Biol. Soc. (EMBC)}.\hskip 1em plus 0.5em
  minus 0.4em\relax IEEE, 2015, pp. 2608--2611.

\bibitem{jing2021ecg}
E.~Jing, H.~Zhang, Z.~Li \emph{et~al.}, ``{ECG} heartbeat classification based
  on an improved {ResNet}-18 model,'' \emph{Comput. Math. Methods Med.}, vol.
  2021, no.~1, p. 6649970, 2021.

\bibitem{yang2025robust}
G.~Yang, S.~Zou, H.~Qin \emph{et~al.}, ``Robust 12-lead {ECG} classification
  with lightweight {ResNet}: An adaptive second-order learning rate
  optimization approach,'' \emph{Electronics}, vol.~14, no.~10, p. 1941, 2025.

\bibitem{wasimuddin2021multiclass}
M.~Wasimuddin, K.~Elleithy \emph{et~al.}, ``Multiclass {ECG} signal analysis
  using global average-based 2-{D} convolutional neural network modeling,''
  \emph{Electronics}, vol.~10, no.~2, p. 170, 2021.

\bibitem{crocker2021inceptiontime}
H.~J. Crocker and A.~W. Costall, ``An {InceptionTime}-inspired convolutional
  neural network to detect cardiac abnormalities in reduced-lead {ECG} data,''
  in \emph{Comput. Cardiol. (CinC)}, vol.~48.\hskip 1em plus 0.5em minus
  0.4em\relax IEEE, 2021, pp. 1--4.

\bibitem{cao2023incepse}
T.~Cao, N.~Tran, L.~Nguyen, H.~Nguyen, and H.~Pham, ``{IncepSE}: Leveraging
  {InceptionTime}'s performance with squeeze and excitation mechanism in {ECG}
  analysis,'' in \emph{Proc. 12th Int. Symp. Inf. Commun. Technol.}, 2023, pp.
  578--584.

\bibitem{zhao2025mak}
C.~Zhao, B.~Lai, Y.~Xu, Y.~Wang, and H.~Dong, ``{MAK-Net}: A multi-scale
  attentive {Kolmogorov--Arnold} network with {BiGRU} for imbalanced {ECG}
  arrhythmia classification,'' \emph{Sensors}, vol.~25, no.~13, p. 3928, 2025.

\bibitem{bodini2021opening}
M.~Bodini, M.~W. Rivolta, and R.~Sassi, ``Opening the black box:
  interpretability of machine learning algorithms in electrocardiography,''
  \emph{Philos. Trans. Roy. Soc. A}, vol. 379, no. 2212, p. 20200253, 2021.

\bibitem{duran2021afraid}
J.~M. Dur{\'a}n and K.~R. Jongsma, ``Who is afraid of black box algorithms? on
  the epistemological and ethical basis of trust in medical {AI},'' \emph{J.
  Med. Ethics}, vol.~47, no.~5, pp. 329--335, 2021.

\bibitem{ronan2024brugada}
R.~Ronan, C.~Tarabanis, L.~Chinitz, and L.~Jankelson, ``{Brugada} {ECG}
  detection with self-supervised {VICReg} pre-training: a novel deep learning
  approach for rare cardiac diseases,'' \emph{medRxiv}, 2024, [Preprint]. doi:
  10.1101/2024.03.29.24305072.

\bibitem{kiyasseh2021clocs}
D.~Kiyasseh, T.~Zhu, and D.~A. Clifton, ``{CLOCS}: Contrastive learning of
  cardiac signals across space, time, and patients,'' in \emph{Proc. Int. Conf.
  Mach. Learn. (ICML)}.\hskip 1em plus 0.5em minus 0.4em\relax PMLR, 2021, pp.
  5606--5615.

\bibitem{mousavi2021ecg}
S.~Mousavi, F.~Afghah, F.~Khadem, and U.~R. Acharya, ``{ECG} language
  processing ({ELP}): A new technique to analyze {ECG} signals,'' \emph{Comput.
  Methods Programs Biomed.}, vol. 202, p. 105959, 2021.

\bibitem{mckeen2025ecg}
K.~McKeen, S.~Masood, A.~Toma, B.~Rubin, and B.~Wang, ``{ECG-FM}: An open
  electrocardiogram foundation model,'' \emph{JAMIA Open}, vol.~8, no.~5, p.
  ooaf122, 2025.

\bibitem{wang2024anyecg}
Y.~Wang, X.~Cao, Y.~Hu \emph{et~al.}, ``{AnyECG}: Foundational models for
  electrocardiogram analysis,'' \emph{arXiv preprint arXiv:2411.17711}, 2024.

\bibitem{na2024guiding}
Y.~Na, M.~Park, Y.~Tae, and S.~Joo, ``Guiding masked representation learning to
  capture spatio-temporal relationship of electrocardiogram,'' \emph{arXiv
  preprint arXiv:2402.09450}, 2024.

\bibitem{shivashankara2024ecg}
K.~K. Shivashankara, A.~M. Shervedani \emph{et~al.}, ``{ECG-Image-Kit}: a
  synthetic image generation toolbox to facilitate deep learning-based
  electrocardiogram digitization,'' \emph{Physiol. Meas.}, vol.~45, no.~5, p.
  055019, 2024.

\bibitem{han2024ecg}
W.~Han, C.~Duan, M.~A. Rosenberg, E.~Liu, and D.~Zhao, ``{ECG-Byte}: A
  tokenizer for end-to-end generative electrocardiogram language modeling,''
  \emph{arXiv preprint arXiv:2412.14373}, 2024.

\bibitem{an2024research}
X.~An, S.~Shi, Q.~Wang, Y.~Yu, and Q.~Liu, ``Research on a lightweight
  arrhythmia classification model based on knowledge distillation for wearable
  single-lead {ECG} monitoring systems,'' \emph{Sensors}, vol.~24, no.~24, p.
  7896, 2024.

\bibitem{spicher2021edge}
N.~Spicher, A.~Klingenberg, V.~Purrucker, and T.~M. Deserno, ``Edge computing
  in {5G} cellular networks for real-time analysis of electrocardiography
  recorded with wearable textile sensors,'' in \emph{Proc. 43rd Annu. Int.
  Conf. IEEE Eng. Med. Biol. Soc. (EMBC)}.\hskip 1em plus 0.5em minus
  0.4em\relax IEEE, 2021, pp. 1735--1739.

\bibitem{najia2025enhanced}
M.~Najia and B.~Faouzi, ``An enhanced hybrid model combining {CNN}, {BiLSTM},
  and attention mechanism for {ECG} segment classification,'' \emph{Biomed.
  Eng. Comput. Biol.}, vol.~16, p. 11795972251341051, 2025.

\bibitem{de2021clinical}
A.~B. De~Luna, M.~Fiol-Sala, A.~Bayes-Genis, and A.~Baranchuk, \emph{Clinical
  electrocardiography: a textbook}.\hskip 1em plus 0.5em minus 0.4em\relax John
  Wiley \& Sons, 2021.

\bibitem{wagner2009aha}
G.~S. Wagner, P.~Macfarlane, H.~Wellens, M.~Josephson, A.~Gorgels, D.~M.
  Mirvis, O.~Pahlm, B.~Surawicz, P.~Kligfield, R.~Childers \emph{et~al.},
  ``Aha/accf/hrs recommendations for the standardization and interpretation of
  the electrocardiogram: part vi: acute ischemia/infarction a scientific
  statement from the american heart association electrocardiography and
  arrhythmias committee, council on clinical cardiology; the american college
  of cardiology foundation; and the heart rhythm society endorsed by the
  international society for computerized electrocardiology,'' \emph{Journal of
  the American College of Cardiology}, vol.~53, no.~11, pp. 1003--1011, 2009.

\bibitem{hammer2025explainable}
A.~Hammer, F.~Haensch, L.~Seliger, H.~Malberg, A.~Linke, and M.~Schmidt,
  ``Explainable and interpretable {AI} visualises self-learned clinically
  relevant {ECG} characteristics of rhythm and morphology paving the way for
  trustworthy diagnostic support,'' \emph{Eur. Heart J.}, vol.~46, no.
  Supplement\_1, 2025.

\bibitem{hampton2024ecg}
J.~Hampton, J.~Hampton, and D.~Adlam, \emph{The ecg made easy e-book: The ecg
  made easy e-book}.\hskip 1em plus 0.5em minus 0.4em\relax Elsevier Health
  Sciences, 2024.

\bibitem{he2019bag}
T.~He, Z.~Zhang, H.~Zhang, Z.~Zhang, J.~Xie, and M.~Li, ``Bag of tricks for
  image classification with convolutional neural networks,'' in \emph{Proc.
  IEEE/CVF Conf. Comput. Vis. Pattern Recognit. (CVPR)}, 2019, pp. 558--567.

\bibitem{wang2017time}
Z.~Wang, W.~Yan, and T.~Oates, ``Time series classification from scratch with
  deep neural networks: A strong baseline,'' in \emph{Proc. Int. Joint Conf.
  Neural Netw. (IJCNN)}.\hskip 1em plus 0.5em minus 0.4em\relax IEEE, 2017, pp.
  1578--1585.

\bibitem{ismail2020inceptiontime}
H.~Ismail~Fawaz, B.~Lucas, G.~Forestier \emph{et~al.}, ``{InceptionTime}:
  Finding {AlexNet} for time series classification,'' \emph{Data Mining Knowl.
  Discov.}, vol.~34, no.~6, pp. 1936--1962, 2020.

\bibitem{loshchilov2017decoupled}
I.~Loshchilov and F.~Hutter, ``Decoupled weight decay regularization,''
  \emph{arXiv preprint arXiv:1711.05101}, 2017.

\bibitem{rajpurkar2022ai}
P.~Rajpurkar, E.~Chen, O.~Banerjee, and E.~J. Topol, ``{AI} in health and
  medicine,'' \emph{Nat. Med.}, vol.~28, no.~1, pp. 31--38, 2022.

\bibitem{kligfield2007recommendations}
P.~Kligfield, L.~S. Gettes, J.~J. Bailey, R.~Childers, B.~J. Deal, E.~W.
  Hancock, G.~Van~Herpen, J.~A. Kors, P.~Macfarlane, D.~M. Mirvis
  \emph{et~al.}, ``Recommendations for the standardization and interpretation
  of the electrocardiogram: part i: the electrocardiogram and its technology a
  scientific statement from the american heart association electrocardiography
  and arrhythmias committee, council on clinical cardiology; the american
  college of cardiology foundation; and the heart rhythm society endorsed by
  the international society for computerized electrocardiology,'' \emph{Journal
  of the American College of Cardiology}, vol.~49, no.~10, pp. 1109--1127,
  2007.

\bibitem{goldberger2023goldberger}
A.~L. Goldberger, Z.~D. Goldberger, and A.~Shvilkin, \emph{Goldberger's
  Clinical Electrocardiography-E-Book: A Simplified Approach}.\hskip 1em plus
  0.5em minus 0.4em\relax Elsevier Health Sciences, 2023.

\bibitem{yang2018independent}
C.~Yang and N.~Tavassolian, ``An independent component analysis approach to
  motion noise cancelation of cardio-mechanical signals,'' \emph{IEEE
  Transactions on Biomedical Engineering}, vol.~66, no.~3, pp. 784--793, 2018.

\end{thebibliography}

\end{document}